\documentclass{article}
\PassOptionsToPackage{table}{xcolor}
\usepackage{amssymb}

\usepackage[preprint]{corl_2025} 

\usepackage{amsmath,amsfonts,bm}

\def\eqref#1{equation~\ref{#1}}

\def\1{\bm{1}}

\DeclareMathAlphabet{\mathsfit}{\encodingdefault}{\sfdefault}{m}{sl}
\SetMathAlphabet{\mathsfit}{bold}{\encodingdefault}{\sfdefault}{bx}{n}

\def\gA{{\mathcal{A}}}

\def\gC{{\mathcal{C}}}

\def\gE{{\mathcal{E}}}

\def\gJ{{\mathcal{J}}}

\def\gM{{\mathcal{M}}}
\def\gN{{\mathcal{N}}}
\def\gO{{\mathcal{O}}}
\def\gP{{\mathcal{P}}}

\def\gS{{\mathcal{S}}}

\def\gU{{\mathcal{U}}}

\usepackage{enumitem}
\usepackage{microtype}
\usepackage{graphicx}
\usepackage{subcaption} 
\usepackage{booktabs} 
\usepackage{lipsum}
\usepackage{hyperref}
\usepackage{xspace}

\newcommand{\framework}{AT-MDP framework\xspace}

\newcommand{\Yes}{\textbf{\textcolor{red}{Yes}}}
\newcommand{\No}{\textcolor{black}{No}}

\newcommand{\superhard}{{4p3e5o}\xspace} 

\usepackage{amsmath}
\usepackage{amssymb}
\usepackage{mathtools}
\usepackage{amsthm}
\usepackage{multirow}
\usepackage{multicol}

\usepackage[capitalize,noabbrev]{cleveref}

\theoremstyle{plain}
\newtheorem{theorem}{Theorem}[section]

\theoremstyle{definition}
\newtheorem{definition}[theorem]{Definition}

\theoremstyle{remark}

\usepackage{float}

\usepackage[textsize=tiny]{todonotes}

\usepackage{listings}

\definecolor{strcolor}{HTML}{1B51A5}
\definecolor{keycolor}{HTML}{278559}
\lstdefinelanguage{json}{
    basicstyle=\ttfamily\footnotesize,            
    numbers=none,                    
    numberstyle=\tiny,               
    stepnumber=1,                    
    numbersep=1pt,                   
    showstringspaces=false,          
    breaklines=true,                 
    frame=shadowbox,                 
    rulesepcolor=\color{gray!30},    
    backgroundcolor=\color{white}, 
    stringstyle=\color{strcolor},      
    keywordstyle=\bfseries\color{keycolor}, 
    morestring=[b]"                  
}

\title{AT-Drone: Benchmarking Adaptive Teaming in Multi-Drone Pursuit}

\author{
    Yang Li$^{1}$,
  Junfan~Chen$^{2}$, 
  Feng~Xue$^{3}$,
  Jiabin~Qiu$^{4}$,
  Wenbin~Li$^{4}$,
  Qingrui~Zhang$^{3}$, 
  \\
  \textbf{Ying~Wen}$^{2\ *}$ ,
  \textbf{Wei~Pan}$^{1}$ \thanks{Corresponding authors. } \\ 
  $^1$ University of Manchester; $^2$ Shanghai Jiao Tong University; \\ $^3$ Sun Yat-sen University; $^4$ Nanjing University; 
}

\begin{document}
\maketitle


\begin{abstract}
Adaptive teaming—the capability of agents to effectively collaborate with unfamiliar teammates without prior coordination—is widely explored in virtual video games but overlooked in real-world multi-robot contexts. 
Yet, such adaptive collaboration is crucial for real-world applications, including border surveillance, search-and-rescue, and counter-terrorism operations.
To address this gap, we introduce AT-Drone, the first dedicated benchmark explicitly designed to facilitate comprehensive training and evaluation of adaptive teaming strategies in multi-drone pursuit scenarios. 
AT-Drone makes the following key contributions:
(1) An adaptable simulation environment configurator that enables intuitive and rapid setup of adaptive teaming multi-drone pursuit tasks, including four predefined pursuit environments.
(2) A streamlined real-world deployment pipeline that seamlessly translates simulation insights into practical drone evaluations using edge devices and Crazyflie drones.
(3) A novel algorithm zoo integrated with a distributed training framework, featuring diverse algorithms explicitly tailored, for the first time, to multi-pursuer and multi-evader settings.
(4) Standardized evaluation protocols with newly designed unseen drone zoos, explicitly designed to rigorously assess the performance of adaptive teaming.
Comprehensive experimental evaluations across four progressively challenging multi-drone pursuit scenarios confirm AT-Drone's effectiveness in advancing adaptive teaming research. 
Real-world drone experiments further validate its practical feasibility and utility for realistic robotic operations.
Videos, code and weights are available at \url{https://sites.google.com/view/at-drone}. 

\end{abstract}



\section{Introduction}

Multi-drone collaboration has become increasingly critical for a variety of real-world applications, including disaster response, border surveillance, and search-and-rescue missions~\cite{chung2011search,ZhangDACOOP2023,queralta2020collaborative}. 
The success of these missions heavily relies on drones' capability to effectively coordinate in real-time within dynamic environments with changing team compositions and unpredictable target behaviors. 
For example, in disaster scenarios, drones may become damaged or depleted during operation, necessitating rapid integration of backup drones to sustain mission effectiveness. Adaptive teaming directly addresses these challenges by enabling drones to dynamically collaborate with previously unseen teammates, significantly enhancing the operational robustness and flexibility of drone fleets.

However, current multi-drone collaboration methodologies predominantly rely on pre-defined coordination mechanisms or extensive prior interactions among drones, thus limiting their adaptability to unexpected or new teammates.
Traditional optimization-based approaches~\cite{shah_multi-agent_2019,janosov_group_2017,zhou_cooperative_2016} and reinforcement learning methods~\cite{ZhangDACOOP2023,chen2024dualcurriculumlearningframework,qi_cascaded_2024,de_souza_decentralized_2021,li_robust_2019,matignon_hysteretic_2007} typically utilize fixed conventions, roles, or communication protocols, hindering their performance when encountering unfamiliar teammates or unpredictable environments.

Conversely, existing adaptive teaming research—such as zero-shot coordination (ZSC)~\cite{hu2020other} and ad-hoc teamwork (AHT)~\cite{stone2010ad}—mainly focuses on simulated environments with discrete action spaces, exemplified by video games such as Overcooked~\cite{li2023cooperative,wang2024zsc}, Hanabi~\cite{hu2020other,anyplay,canaan2022generating,bard2020hanabi}, and Predator-Prey~\cite{barrett2011empirical,papoudakis2021agent}. 
Recent advancements like NAHT~\cite{wang2024n} extend these paradigms to multiple learners but remain restricted within discrete-action domains such as the SMAC benchmark~\cite{samvelyan2019starcraft}, limiting their real-world applicability.

\begin{table*}[t]
    \centering
    \vspace{-3mm}
    \caption{Comparison of related work. Grey rows represent  literature related to multi-drone pursuit, while pink rows highlight adaptive teaming studies from the machine learning field. ``AT w/o TM'' and ``AT w/ TM'' denote adaptive teaming without and with teammate modelling, respectively.}
    \resizebox{\linewidth}{!}{
    \begin{tabular}{c|c|c|c|c|c|c|c|c}
        \toprule
        \multirow{2}{*}{\textbf{Related Work}} & \multicolumn{4}{c|}{\textbf{Problem Setting}}& \multicolumn{2}{c|}{\textbf{Task}} & \multicolumn{2}{c}{\textbf{Method}} \\
        \cline{2-9}
        & \textbf{\# Learner} & \textbf{\# Unseen} & \textbf{\# Opponent} &\textbf{Action Space} & \textbf{Main Related Task} & \textbf{Real-world?} & \textbf{AT w/o TM?} & \textbf{AT w/ TM?} \\
        \midrule 
        \rowcolor{gray!10} 
        Voronoi Partitions~\cite{zhou_cooperative_2016} & Multi & 0 & 1  & Continuous &  Pursuit–evasion Game & \No & \No & \No \\ 
        \cline{1-9}
        \rowcolor{gray!10} 
        Bio-pursuit~\cite{janosov_group_2017}  & Multi & 0 & Multi & Continuous &  Prey–predator Game & \No & \No & \No \\
        \cline{1-9}
        \rowcolor{gray!10} 
        Uncertainty-pursuit~\cite{shah_multi-agent_2019} & Multi & 0 & 1 & Continuous & Pursuit–evasion Game & \No & \No & \No \\ 
        \cline{1-9}
        \rowcolor{gray!10}
        M3DDPG~\cite{li_robust_2019} & Multi & 0 & 1 & Continuous &  Prey–predator Game & \No & \No & \No \\ 
        \cline{1-9}
        \rowcolor{gray!10}
        Pursuit-TD3\cite{de_souza_decentralized_2021} & Multi & 0 & 1 & Continuous &  \textbf{\textcolor{blue}{Multi-drone Pursuit}} & \Yes & \No & \No \\ 
        \cline{1-9}
        \rowcolor{gray!10} 
        DACOOP-A\cite{ZhangDACOOP2023} & Multi & 0 & 1 & Discrete &  \textbf{\textcolor{blue}{Multi-drone Pursuit}} & \Yes & \No & \No \\ 
        \cline{1-9}
        \rowcolor{gray!10} 
        GM-TD3~\cite{zhang2024multi}  & Multi & 0 & 1 & Continuous & Prey–predator Game & \No & \No & \No \\ 
        \cline{1-9}
        \rowcolor{gray!10} 
        DualCL~\cite{chen2024dualcurriculumlearningframework} & Multi & 0 & 1 & Continuous & \textbf{\textcolor{blue}{Multi-drone Pursuit}} & \No & \No & \No \\ 
        \cline{1-9}
        \rowcolor{gray!10}
        HOLA-Drone~\cite{hola-drone}  &  1 & Multi & Multi & Continuous & \textbf{\textcolor{blue}{Multi-drone Pursuit}} & \Yes & \Yes  & \No \\ 
        \midrule
        \rowcolor{pink!30} 
        Other-play~\cite{hu2020other} & 1 & 1 & 0 &  Discrete & Lever Game; Hanabi &  \No & \Yes  & \No \\ 
        \cline{1-9}
        \rowcolor{pink!30} 
        Overcooked-AI ~\cite{carroll2019utility} & 1 & 1 & 0 &  Discrete & Overcooked &  \No & \Yes  & \No \\ 
        \cline{1-9}
        \rowcolor{pink!30} 
        TrajDi~\cite{TrajDi}  & 1 & 1 & 0 &  Discrete & Overcooked &  \No & \Yes  & \No \\ 
        \cline{1-9}
        \rowcolor{pink!30} 
        MEP~\cite{MEP}  & 1 & 1 & 0 &  Discrete & Overcooked &  \No & \Yes  & \No \\ 
        \cline{1-9}
        \rowcolor{pink!30} 
        LIPO~\cite{charakorn2023generating}  & 1 & 1 & 0 &  Discrete & Overcooked &  \No & \Yes  & \No \\ 
        \cline{1-9}
        \rowcolor{pink!30} 
        HSP~\cite{yu2023learning}  & 1 & 1 & 0 &  Discrete & Overcooked &  \No & \Yes  & \No \\ 
        \cline{1-9}
        \rowcolor{pink!30} 
        COLE~\cite{li2024tackling}  & 1 & 1 & 0 &  Discrete & Overcooked &  \No & \Yes  & \No \\ 
        \cline{1-9}
        \rowcolor{pink!30} 
        ZSC-Eval~\cite{wang2024zsc}  & 1 & 1 & 0 &  Discrete & Overcooked &  \No & \Yes  & \No \\ 
        \cline{1-9}
       \rowcolor{pink!30}  
       PLASTIC~\cite{barrett2017making}  &  1 & Multi & Multi & Discrete & Prey-predator Game &  \No & \No & \Yes \\ 
       \cline{1-9} 
        \rowcolor{pink!30} 
        AATeam~\cite{chen2020aateam}  & 1 & 1 & 2 &  Discrete & Half Field Offense &  \No & \No & \Yes  \\  
        \cline{1-9} 
        \rowcolor{pink!30} 
        LIAM~\cite{papoudakis2021agent}  &  1 & Multi & 
        Multi & Discrete & LBF; Prey-predator Game  &  \No & \No & \Yes   \\ 
        \cline{1-9} 
       \rowcolor{pink!30}  
       GPL~\cite{rahman2021towards}  &  1 & Multi & Multi & Discrete & LBF; Wolfpack; FortAttack &  \No & \No & \Yes   \\ 
       \cline{1-9} 
       \rowcolor{pink!30}   
       CIAO~\cite{jianhong2024oaht} &  1 & Multi & Multi & Discrete &LBF; Wolfpack &  \No & \No & \Yes \\ 
       \cline{1-9} 
       \rowcolor{pink!30}   
       NAHT~\cite{wang2024n}  & Multi & Multi & Multi & Discrete & StarCraft; MPE &  \No & \No & \Yes \\
         \bottomrule
    \end{tabular}
    }
    \label{tab:review}
    \vspace{-5mm}
\end{table*}

As summarized in Table~\ref{tab:review}, there is a clear lack of benchmarks tailored specifically for studying adaptive teaming in complex, real-world scenarios involving multi-drone collaboration. To address this critical gap, we introduce AT-Drone, the first unified benchmark explicitly designed to integrate adaptive teaming methods from machine learning into practical multi-drone robotics applications. AT-Drone provides a comprehensive and standardized training and evaluation framework, facilitating rapid assessment of adaptive teaming algorithms and effectively bridging theoretical innovations with real-world robotic deployments.

AT-Drone consists of four main components designed to thoroughly study adaptive teaming in multi-drone pursuit tasks: \textbf{1. A customizable simulation environment:} AT-Drone includes four progressively challenging multi-drone pursuit environments, systematically varying in obstacle complexity, evader numbers, and task difficulty, enabling rigorous testing of adaptive strategies across diverse operational conditions. \textbf{2. Streamlined real-world deployment pipeline:} AT-Drone integrates practical deployment pipelines that leverage motion capture systems and edge devices (such as Nvidia Jetson Orin Nano). Real-world experiments with Crazyflie drones validate the benchmark's fidelity and reflect its potential for advancing more complex real-world applications in the future. 
\textbf{3. A novel algorithm zoo:} AT-Drone introduces a distributed training infrastructure comprising seven adaptive teaming algorithms adapted and extended from discrete video game environments. To the best of our knowledge, this benchmark represents the first systematic exploration of adaptive teaming strategies tailored to multi-pursuit multi-evader drone pursuit scenarios. \textbf{4. Standardized evaluation protocols:} AT-Drone provides three distinct ``unseen drone zoo'' configurations, each demanding unique adaptive collaboration strategies, alongside four specialized evaluation metrics designed to systematically assess algorithm adaptability and robustness.

\vspace{-2mm}
\section{Related Work}
\vspace{-3mm}

As summarised in Table~\ref{tab:review}, we provide a detailed comparison of related methods across key dimensions, including problem formulation, task scope, and methodological approaches, highlighting the unique positioning of our benchmark within the literature.

\textbf{Multi-agent pursuit-evasion. } 
Multi-agent pursuit-evasion is closely related to the multi-drone pursuit task. 
Most existing methods rely on pre-coordinated strategies specifically designed for particular pursuit-evasion scenarios.
Traditional approaches often rely on heuristic~\cite{janosov_group_2017} or optimisation-based strategies~\cite{zhou_cooperative_2016,shah_multi-agent_2019}. 
In recent years, deep reinforcement learning (DRL) has been widely adopted for pre-coordinated multi-drone pursuit tasks. M3DDPG~\cite{li_robust_2019} and GM-TD3~\cite{zhang2024multi} extend standard DRL algorithms, such as TD3~\cite{TD3} and DDPG~\cite{DDPG}, specifically for multi-agent pursuit in simulated environments. Pursuit-TD3~\cite{de_souza_decentralized_2021} applies the TD3 algorithm to pursue a target with multiple homogeneous agents, which is validated through both simulations and real-world drone demonstrations. \citet{ZhangDACOOP2023} introduces DACOOP-A, a cooperative pursuit algorithm that enhances reinforcement learning with artificial potential fields and attention mechanisms, which is evaluated in real-world drone systems. DualCL~\cite{chen2024dualcurriculumlearningframework} addresses multi-UAV pursuit-evasion in diverse environments and demonstrates zero-shot transfer capabilities to unseen scenarios, though only in simulation.
The most recent work, HOLA-Drone~\cite{hola-drone}, claims to be the first ZSC framework for multi-drone pursuit. However, it is limited to controlling a single learner, restricting its applicability to broader multi-agent settings. 

\textbf{Adaptive Teaming. }The adaptive teaming paradigm can be broadly categorised into two aspects: adaptive teaming without teammate modelling (AT w/o TM) and adaptive teaming with teammate modelling (AT w/ TM), which correspond to the zero-shot coordination (ZSC) and ad-hoc teamwork (AHT) problems in the machine learning community, respectively.
AT w/o TM focuses on enabling agents to coordinate with unseen teammates without explicitly modelling their behaviours. Other-Play~\cite{hu2020other} introduces an approach that leverages symmetries in the environment to train robust coordination policies, applied to discrete-action tasks like the Lever Game and Hanabi. Similarly, methods such as Overcooked-AI~\cite{carroll2019utility}, TrajDi~\cite{TrajDi}, MEP~\cite{MEP}, LIPO~\cite{charakorn2023generating}, and ZSC-Eval~\cite{wang2024zsc} study collaborative behaviours in Overcooked, where agents learn generalisable coordination strategies with diverse unseen partners. While these approaches demonstrate promising results, they are limited to single-learner frameworks in simplified, discrete-action domains like Overcooked and Hanabi. They lack scalability to multi-agent settings, continuous action spaces, and the complexities of real-world applications.
AT w/ TM, on the other hand, explicitly models the behaviour of unseen teammates to facilitate effective collaboration. Early methods like PLASTIC~\cite{barrett2017making} reuse knowledge from previous teammates or expert input to adapt to new teammates efficiently. AaTeam~\cite{chen2020aateam} introduces attention-based neural networks to dynamically process and respond to teammates’ behaviours in real-time.
More advanced approaches, such as LIAM~\cite{papoudakis2021agent}, employ encoder-decoder architectures to model teammates using local information from the controlled agent. GPL~\cite{rahman2021towards} and CIAO~\cite{jianhong2024oaht} leverage GNNs to address the challenges of dynamic team sizes in AHT. Extending from the AHT settings,
NAHT~\cite{wang2024n} enables multiple learners to collaborate and interact with diverse unseen partners in N-agent scenarios.
Despite their progress, these methods remain confined to discrete action spaces and simulated benchmarks, limiting their applicability to real-world, continuous-action tasks. 

\begin{figure*}[!ht]
    \centering
    \includegraphics[width=\linewidth]{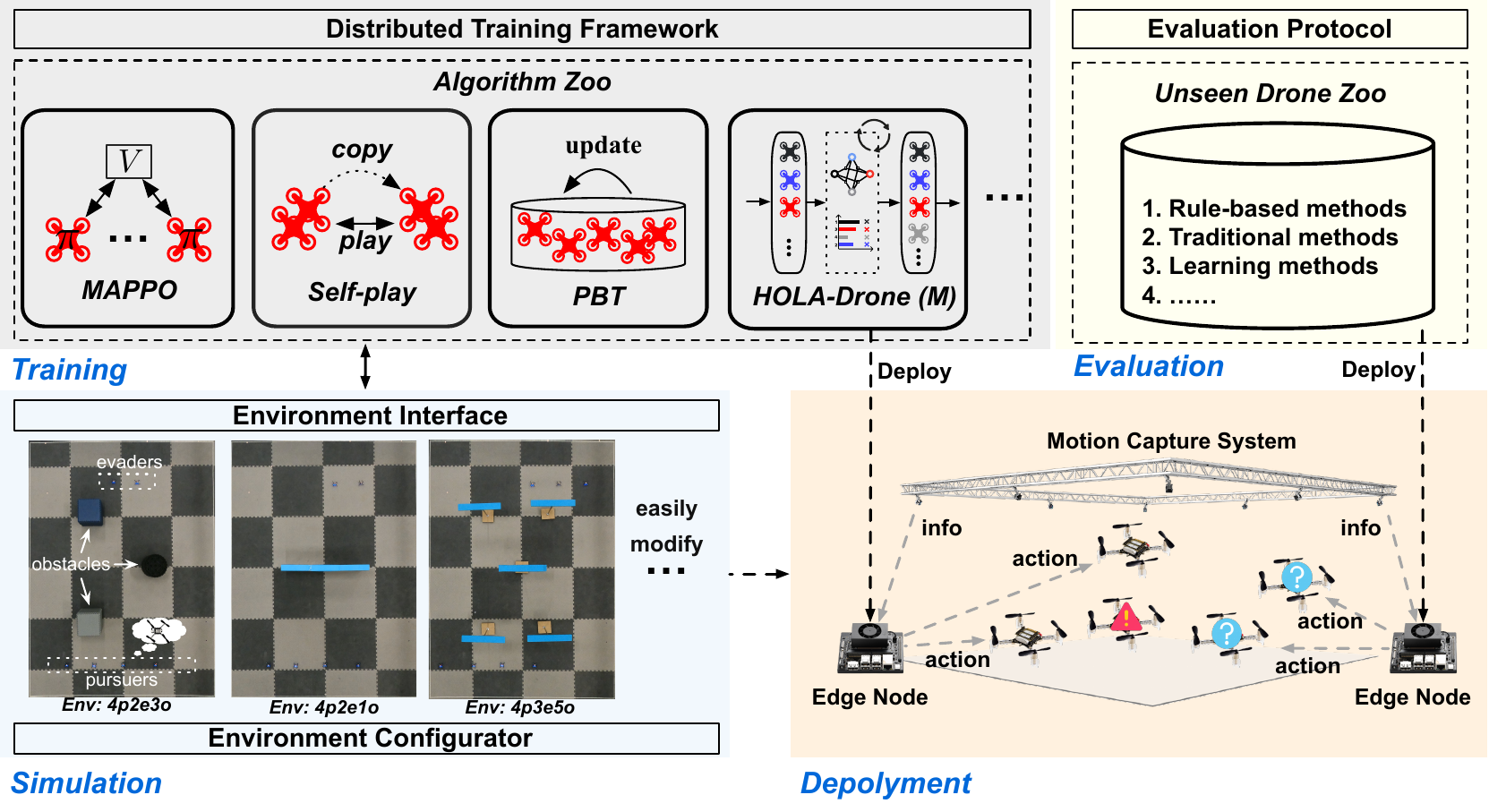}
    \caption{Overview of the AT-Drone Benchmark, comprising four key components: (I) a customizable simulation environment featuring varied multi-drone pursuit tasks with adjustable complexity; (II) a streamlined real-world deployment pipeline employing motion capture systems and edge devices to facilitate realistic drone validation; (III) a distributed training framework equipped with diverse adaptive teaming algorithms for multi-drone pursuit task; and (IV) standardized evaluation protocols, leveraging diverse unseen teammate configurations to rigorously evaluate adaptive teaming performance and robustness across distinct strategies.
    }
    \vspace{-5mm}
    \label{fig:system}
\end{figure*}

\section{The Benchmark: AT-Drone}
\label{sec_mdpat}

\subsection{Problem Formulation}
\label{sec:method_1}
\begin{definition}[Adaptive Teaming in Multi-Drone Pursuit]
Adaptive teaming in multi-drone pursuit involves training a set of $N \in \{1, 2, \dots\}$ drone agents, referred to as learners, to dynamically coordinate with $M \in \{1, 2, \dots\}$ previously unseen partners. The objective is to pursue $K \in \{1, 2, \dots\}$ targets without collisions, optimizing the overall return.
\end{definition}

Let $\gC$ represent the cooperative team, comprising $N$ learners and $M$ uncontrolled teammates. The set of uncontrolled teammates is denoted by $\gU$. In the multi-drone pursuit task, there exists a set of opponents, denoted as $\gE$.
Adaptive teaming can be effectively modeled as an extended \textbf{A}daptive \textbf{T}eaming \textbf{Dec}entralized \textbf{P}artially \textbf{O}bservable \textbf{M}arkov \textbf{D}ecision \textbf{P}rocess (AT-Dec-POMDP). AT-Dec-POMDP is defined by the tuple $(\gS, \gC, \gA, \gP, r, \gO, \gamma, T)$, where:
where $\gS$ is the joint state space; $\gC$ denotes the set of cooperative agents, consisting of learners ($\gN$) and uncontrolled teammates ($\gM$), where $\gM$ is sampled according to  $\gP_\text{u}(\gM | \gU)$ from the complete set of uncontrolled teammates $\gU$; $\gA = \times_{j=1}^C \gA^j$ is the joint action space, where $C = N + M$ is the team size; $\gP(s’|s, a)$ is the transition probability function, representing the probability of transitioning to state $s’ \in \gS$ given the current state $s \in \gS$ and joint action $a \in \gA$; $r(s, a)$ is the reward function, representing the team’s reward in state $s$ after taking action $a$; $\gO$ is the joint observation space, with $\gO(o|s)$ describing the probability of generating observation $o$ given state $s$; $\gamma \in [0, 1]$ is the discount factor; and $T$ is the task horizon.

Additionally, we denote the policy of agent $j$ as $\pi^j$, through which the agent selects an action $a_t^j \in \gA^j$, and the policies of the $N$ learners ($\pi^i$, for $i \in \gN$) are learnable; we consider two approaches for defining these policies: with and without teammate modeling.
Adaptive teaming without teammate modeling is closely related to the zero-shot coordination problem~\cite{hu2020other,carroll2019utility}, where learners could coordinate with unseen teammates. Specifically, the policy for a learner $i$ is represented as $\pi^i(a_t^i \mid \tau_t^i)$, where $\tau_t^i$ denotes the learner's observation history up to time $t$.
On the other hand, adaptive teaming with teammate modeling aligns closely with the ad-hoc teamwork paradigm~\cite{stone2010ad}, where agents could coordinate with previously unknown teammates by explicitly modeling their behavior and characteristics.
In this case, the policy is defined as $\pi^i(a_t^i \mid \tau_t^i, f(\tau_t^i))$.
The joint action $a_t = (a_t^1, \dots, a_t^C)$ determines the next state $s_{t+1} \sim \gP(s_{t+1} \mid s_t, a_t)$, and all agents receive a shared reward $r(s_t, a_t)$.
The goal of adaptive teaming is to learn policies $\{\pi^i\}_{i \in \gN}$ that maximize the expected discounted return:
$
    \gJ = \mathbb{E}[R(\tau)] = \mathbb{E}\left[\sum_{t=0}^T \gamma^t r(s_t, a_t)\right],
    \label{eq:obj}
$ where $\tau$ denotes the joint trajectory.

\subsection{Simulation and Deployment}
\label{sec:method_2}

\textbf{Simulation.} The simulation module provides a highly customizable and intuitive framework through the environment configurator and the Gymnasium-based environment interface, enabling efficient setup and execution of multi-drone pursuit scenarios. 
The environment configurator organizes simulation parameters into three distinct categories for easy customization: players, site, and task (see Fig.~\ref{fig:env_json} in Appendix~\ref{appendix:env_config}). The {players} category specifies the numbers, velocities, and characteristics of learners, unseen teammates, and evaders, and optionally incorporates an unseen drone zoo to simulate varied adaptive teaming conditions. The {site} category allows users to adjust the simulation environment’s physical properties, including map dimensions and obstacle configurations, enabling a wide range of scenario complexities. The {task} category defines rules and objectives unique to each pursuit scenario, determining conditions for success and interaction dynamics.

Leveraging the configurator, we systematically design four progressive multi-drone pursuit environments in the benchmark, named \texttt{4p2e3o}, \texttt{4p2e1o}, \texttt{4p2e5o}, and \texttt{4p3e5o}, indicating the number of pursuers (p), evaders (e), and obstacles (o). Screenshots of these real-world environments are provided in Fig.~\ref{fig:app_env} in the Appendix. Evaders spawn within a specified region measuring 3.2m wide and 0.6m high, with pursuers spawning in a similar area. Obstacle configurations vary significantly across environments to introduce different complexity levels. Environment \texttt{4p2e3o}, considered easy, contains three distributed obstacles (two cubes and one cylinder), providing ample pursuit space. Although environment \texttt{4p2e1o} has only a single central obstacle, it poses a slightly higher difficulty due to evaders having greater freedom of movement. Environments featuring five obstacles (\texttt{4p2e5o} and \texttt{4p3e5o}) are challenging, requiring sophisticated maneuvering due to densely packed obstacles restricting drone movements. Specifically, \texttt{4p2e5o} is categorized as hard, while \texttt{4p3e5o} is identified as the most challenging scenario (superhard), necessitating advanced coordination strategies for successful adaptive teaming.

In these environments, each agent's observation includes: (1) the relative position and bearing to each evader within its perception range, (2) the distance and angle to the nearest obstacle, and (3) the relative position and bearing of nearby teammates. Observations are structured as normalized continuous vectors, with masking applied to represent occluded entities clearly. The action space is continuous, defined within the range \([-1, 1]\), directly corresponding to angular steering adjustments that control the drone's directional rotation during pursuit. The reward function incentivizes efficient pursuit behaviors through positive rewards for capturing evaders and additional shaped rewards proportional to the agents' proximity improvement towards targets. Safety is enforced via proximity-based penalties for approaching too close to obstacles or teammates, effectively discouraging risky behaviors and promoting cooperative, safe navigation.

\textbf{Deployment.} As shown on the right side of Fig.~\ref{fig:system}, the AT-Drone benchmark supports real-world deployment within a 3.6m$\times$~5m area by seamlessly integrating edge computing nodes—such as the Nvidia Jetson Orin Nano and personal laptops—with Crazyflie drones.
Specifically, we utilize the FZMotion system to perform real-time position tracking, transmitting positional data in point cloud format to Crazyswarm, where it is processed and fed into the decision-making policies.
Policies for both the adaptive learners and unseen drone partners (sampled from the unseen drone pool) run on edge computing nodes that serve as inference engines.
The Crazyswarm platform and the adaptive teaming policies are deployed separately across two edge devices: a Lenovo ThinkPad T590 laptop and a Jetson Orin Nano. These nodes handle the reception of drone position data from the motion capture system, execute the adaptive teaming algorithms, and transmit control commands to the Crazyflie drones via Crazyradio PA.
Upon receiving control signals, the Crazyflie drones execute the maneuvers using their onboard Mellinger controller, ensuring accurate and responsive trajectory tracking.
Overall, this real-world deployment setup allows us to directly evaluate the applicability of learned policies on physical drone systems, effectively bridging the gap between simulation and real-world application.

\vspace{-2mm}
\subsection{Training and Evaluation}
\label{sec:method_3}
\vspace{-3mm}

\textbf{Training.} As illustrated in Fig.~\ref{fig:system}, the AT-Drone framework employs a distributed training architecture leveraging multiple parallel environments to efficiently scale the learning processes. To the best of our knowledge, this is the first study specifically targeting adaptive teaming strategies in multi-pursuer multi-evader drone pursuit scenarios, highlighting the novelty and importance of establishing a comprehensive algorithm zoo for rigorous benchmarking and broader community adoption.

To further boost zero-shot coordination capabilities, AT-Drone employs self-play and PBT strategies. Self-play enables drones to iteratively refine policies via competitive interactions against progressively updated versions of themselves, significantly improving their adaptability and coordination. Simultaneously, PBT~\cite{carroll2019utility} facilitates extensive exploration and efficient knowledge sharing among diverse model populations, thus enhancing generalization to dynamic pursuit tasks. Both self-play and PBT are implemented using Independent PPO (IPPO)~\cite{PPO}, providing a robust, scalable training environment ideal for real-world multi-drone collaboration.

Recently, the HOLA-Drone method~\cite{hola-drone} introduces a hypergraphical-form game to model single-learner scenarios involving interactions with multiple unseen teammates. However, this method faces significant limitations when extended to more complex multi-learner scenarios common in multi-drone pursuit tasks. To overcome these challenges, we propose \textbf{HOLA-Drone V2}, an enhanced and generalized approach. Our key improvements include the construction of a {preference hypergraph} to explicitly identify and retain optimal teammate interactions, combined with a novel {max-min preference oracle}. This oracle systematically identifies challenging teammate subsets and iteratively optimizes drone strategies to robustly handle these scenarios. By dynamically adjusting the learner strategy set, our method significantly improves adaptability and coordination effectiveness in realistic multi-agent environments. Comprehensive algorithmic details, formal definitions, and implementations are provided in Appendix~\ref{appendix:opt}.

In addition to zero-shot coordination methods, we introduce an ad-hoc teamwork algorithm, \textbf{NAHT-D} (NAHT for Drones), based on the recent NAHT framework~\cite{wang2024n}. Unlike the original NAHT method for discrete-action tasks (e.g., SMAC), NAHT-D is adapted for continuous-action drone scenarios, crucial for realistic maneuvering and pursuit tasks. NAHT-D efficiently models unseen drone teammates by integrating a specialized teammate-modeling network into MAPPO~\cite{yu2022surprising}. The teammate-modeling network employs an autoencoder, taking the past $k$ steps of observations and actions to reconstruct teammates' current action distributions. For continuous action spaces, we use KL divergence as the reconstruction loss. The encoder's output embedding captures teammate behaviors and is combined with the agent’s current observations as input to the policy network.
For detailed algorithmic implementation, please refer to Appendix~\ref{appendix:naht-d}.

\textbf{Evaluation.} To rigorously evaluate adaptive teaming strategies within multi-drone pursuit scenarios, we design a structured evaluation protocol consisting of multiple distinct unseen drone zoos and clearly defined evaluation metrics.

We construct a set of \textbf{unseen drone zoos} to introduce diverse and challenging teammate behaviors, spanning rule-based, bio-inspired, and learning-based methods. Specifically, these zoos include: (1) the {Greedy Drone}, employing a straightforward pursuit strategy focused on the nearest evader while dynamically avoiding collisions; (2) the {VICSEK Drone}, inspired by swarm behaviors, optimizing collective drone movement for pursuit and collision avoidance; and (3) the {Self-Play Drones}, trained using randomized IPPO-based self-play to generate diverse and unpredictable coordination behaviors. To ensure thorough evaluation, we define three distinct configurations of unseen drone partners: \textbf{Unseen Zoo 1}: Greedy drones only, highlighting direct pursuit behaviors.
\textbf{Unseen Zoo 2}: Includes two IPPO self-play drone policies demonstrating different coordination skill levels—one highly coordinated (70\% success) and one less coordinated (54\% success)—introducing variability in teaming performance.
\textbf{Unseen Zoo 3}: Combines all drones from the previous zoos, randomly selecting partners each episode to maximize behavioral diversity and unpredictability.
Detailed descriptions and implementations of these drone behaviors are provided in Appendix~\ref{appendix:unseen_zoo}.

We adopt four quantitative metrics to systematically evaluate adaptive teaming performance:
\textbf{Success Rate (SUC)}: Percentage of episodes successfully completed, defined by capturing both evaders within 0.2 meters.
\textbf{Collision Rate (COL)}: Frequency of collisions between drones (threshold of 0.2 meters) or drones and obstacles (threshold of 0.1 meters), assessing operational safety.
\textbf{Average Success Timesteps (AST)}: Mean number of timesteps required for successful task completion, indicating pursuit efficiency.
\textbf{Average Reward (REW)}: Overall quality and efficiency of agent performance across episodes.


\begin{figure}[t]
    \centering
    \begin{minipage}[t]{\textwidth}
        \centering
        \includegraphics[width=\textwidth]{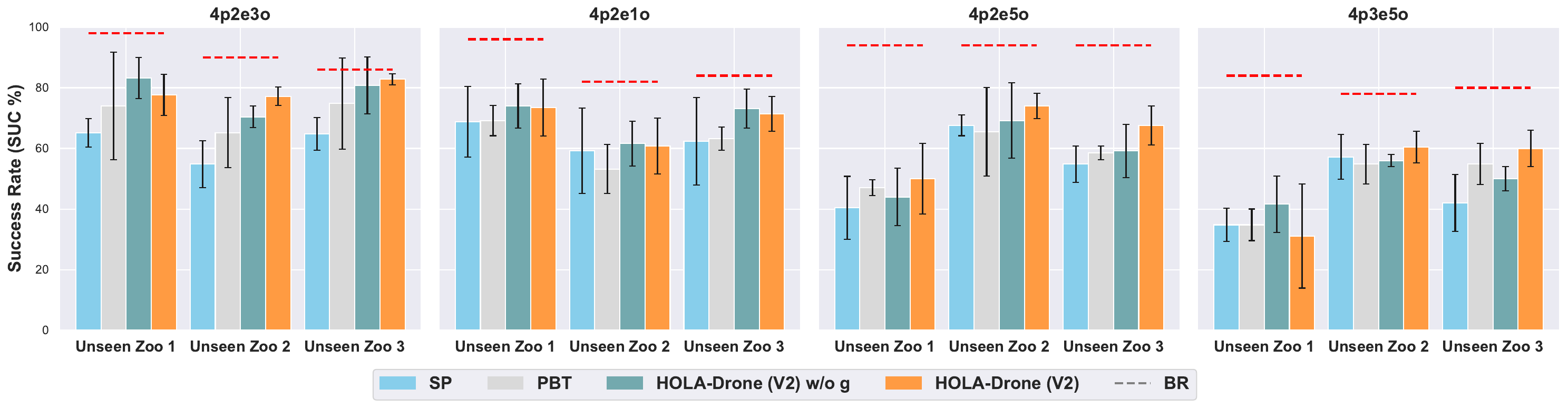}
        \caption{Success rate (SUC) across different difficulty levels for adaptive teaming without teammate modelling. Red dotted lines denote best-response baselines specifically trained on the given unseen teammate zoo.}
        \label{fig:exp1}
    \end{minipage}

    \vspace{1em} 

    \begin{minipage}[t]{\textwidth}
        \centering
        \resizebox{\textwidth}{!}{%
\begin{tabular}{c|c|c|c|c|c|c|c|c|c|c|c|c|c}
\toprule
\multirow{2}{*}{\textbf{ENV}} & \multirow{2}{*}{\textbf{Metrics}}  & \multicolumn{4}{c|}{\textbf{Unseen Zoo 1}} & \multicolumn{4}{c|}{\textbf{Unseen Zoo 2}} & \multicolumn{4}{c}{\textbf{Unseen Zoo 3}} \\
\cline{3-14}
& & \textbf{SP} & \textbf{PBT} & \textbf{H-D w/o g}  & \textbf{H-D }& \textbf{SP} & \textbf{PBT} & \textbf{H-D w/o g}  & \textbf{H-D } & \textbf{SP} & \textbf{PBT} & \textbf{H-D w/o g}  & \textbf{H-D } \\
\midrule
\multirow{6}{*}{\textbf{4p2e3o}} & \multirow{2}{*}{\textbf{COL}$\downarrow$}
&32.40&25.60&\textbf{16.40}&22.33&39.20&34.00&28.80&\textbf{22.40}&31.60&24.00&18.80&\textbf{16.80}\\
&&$\pm$4.34&$\pm$15.19&$\pm$7.27&$\pm$6.81&$\pm$7.35&$\pm$12.10&$\pm$4.15&$\pm$3.58&$\pm$4.90&$\pm$14.38&$\pm$8.79&$\pm$2.28\\
\cline{2-14}
&\multirow{2}{*}{\textbf{AST}$\downarrow$}
&313.48&303.21&260.42&\textbf{259.17}&380.28&376.14&329.24&\textbf{314.97}&380.85&328.96&308.05&\textbf{306.35}\\
&&$\pm$51.91&$\pm$86.38&$\pm$20.68&$\pm$34.28&$\pm$34.65&$\pm$68.19&$\pm$23.35&$\pm$27.68&$\pm$34.36&$\pm$85.25&$\pm$21.18&$\pm$25.86\\
\cline{2-14}
&\multirow{2}{*}{\textbf{REW}$\uparrow$}
&123.60&132.34&\textbf{141.51}&138.28&114.00&121.98&130.38&\textbf{135.71}&126.15&133.11&139.91&\textbf{143.55}\\
&&$\pm$4.68&$\pm$22.68&$\pm$5.75&$\pm$7.95&$\pm$9.07&$\pm$10.13&$\pm$4.78&$\pm$3.46&$\pm$4.68&$\pm$13.39&$\pm$9.48&$\pm$1.29\\
\midrule

\multirow{6}{*}{\textbf{4p2e1o}} & \multirow{2}{*}{\textbf{COL}$\downarrow$}
&30.80&30.40&26.00&\textbf{19.20}&36.80&43.20&36.40&\textbf{32.00}&34.80&35.60&25.60&\textbf{23.60}\\
&&$\pm$12.03&$\pm$5.22&$\pm$7.35&$\pm$7.56&$\pm$15.59&$\pm$10.73&$\pm$8.29&$\pm$5.83&$\pm$14.52&$\pm$5.40&$\pm$6.07&$\pm$6.54\\
\cline{2-14}
&\multirow{2}{*}{\textbf{AST}$\downarrow$}
&352.24&317.55&\textbf{279.43}&298.93&446.67&395.92&383.39&\textbf{353.35}&360.08&375.41&335.27&\textbf{313.60}\\
&&$\pm$25.07&$\pm$30.08&$\pm$29.16&$\pm$29.19&$\pm$51.69&$\pm$72.36&$\pm$43.82&$\pm$35.29&$\pm$22.18&$\pm$48.52&$\pm$36.29&$\pm$34.18\\
\cline{2-14}
&\multirow{2}{*}{\textbf{REW}$\uparrow$}
&122.05&126.73&129.97&\textbf{138.24}&112.66&109.06&114.00&\textbf{120.83}&118.73&120.00&128.76&\textbf{131.50}\\
&&$\pm$9.92&$\pm$6.91&$\pm$9.73&$\pm$6.28&$\pm$11.94&$\pm$13.42&$\pm$10.64&$\pm$6.84&$\pm$13.22&$\pm$5.91&$\pm$8.15&$\pm$5.76\\

\midrule
\multirow{6}{*}{\textbf{4p2e5o}} & \multirow{2}{*}{\textbf{COL}$\downarrow$}
&59.60&53.00&56.00&\textbf{49.86}&31.20&33.50&30.00&\textbf{25.20}&44.00&40.50&39.20&\textbf{32.00}\\
&&$\pm$10.39&$\pm$2.61&$\pm$9.49&$\pm$11.62&$\pm$3.42&$\pm$14.74&$\pm$12.41&$\pm$3.35&$\pm$5.74&$\pm$2.45&$\pm$7.56&$\pm$6.16\\
\cline{2-14}
&\multirow{2}{*}{\textbf{AST}$\downarrow$}
&333.96&\textbf{313.53}&345.89&331.98&287.18&348.23&321.41&\textbf{281.45}&294.41&340.94&332.19&\textbf{313.79}\\
&&$\pm$51.81&$\pm$34.83&$\pm$45.29&$\pm$80.30&$\pm$55.20&$\pm$18.91&$\pm$83.02&$\pm$40.77&$\pm$34.95&$\pm$38.79&$\pm$34.72&$\pm$26.78\\
\cline{2-14}
&\multirow{2}{*}{\textbf{REW}$\uparrow$}
&90.16&93.16&86.30&\textbf{99.31}&120.48&124.92&122.01&\textbf{128.45}&107.56&107.08&104.57&\textbf{119.93}\\
&&$\pm$12.00&$\pm$7.47&$\pm$14.37&$\pm$17.70&$\pm$8.33&$\pm$19.11&$\pm$12.34&$\pm$3.79&$\pm$5.44&$\pm$5.92&$\pm$14.15&$\pm$6.11\\

\midrule

\multirow{6}{*}{\textbf{4p3e5o}} & \multirow{2}{*}{\textbf{COL}$\downarrow$}
&62.80&64.80&\textbf{58.00}&67.57&40.40&38.00&40.00&\textbf{36.40}&55.60&41.60&45.60&\textbf{38.40}\\
&&$\pm$5.02&$\pm$4.60&$\pm$8.72&$\pm$15.85&$\pm$5.83&$\pm$8.05&$\pm$2.45&$\pm$4.34&$\pm$8.65&$\pm$4.77&$\pm$3.29&$\pm$5.18\\
\cline{2-14}
&\multirow{2}{*}{\textbf{AST}$\downarrow$}
&431.44&509.56&\textbf{418.51}&459.55&446.07&555.10&482.64&\textbf{407.85}&425.94&510.96&456.98&\textbf{416.91}\\
&&$\pm$11.17&$\pm$67.14&$\pm$41.49&$\pm$91.71&$\pm$49.11&$\pm$38.97&$\pm$82.82&$\pm$63.06&$\pm$38.53&$\pm$60.10&$\pm$52.86&$\pm$76.21\\
\cline{2-14}
&\multirow{2}{*}{\textbf{REW}$\uparrow$}
&\textbf{141.98}&131.29&136.36&116.57&182.04&187.32&185.98&\textbf{196.07}&146.74&172.06&159.60&\textbf{174.59}\\
&&$\pm$15.98&$\pm$9.66&$\pm$19.43&$\pm$43.94&$\pm$8.97&$\pm$7.17&$\pm$4.43&$\pm$9.91&$\pm$17.01&$\pm$11.93&$\pm$1.74&$\pm$18.68\\

\bottomrule
\end{tabular}%
}
\captionof{table}{Performance comparison of adaptive teaming without teammate modeling across environments with varying difficulties. H-D denotes HOLA-Drone (V2).}
        \label{tab:exp}
    \end{minipage}
    \vspace{-7mm}
\end{figure}

\section{Experiment}
\label{sec:exp}

In this section, we assess baseline methods from our algorithm zoo to validate their practical effectiveness in multi-drone pursuit tasks across four progressively challenging environments: 4p2e3o, 4p2e1o, 4p2e5o, and 4p3e5o. 
The experiments are organized into two primary components: (1) adaptive teaming without explicit teammate modeling, and (2) adaptive teaming incorporating teammate modeling. Each subsection outlines the experimental setups, highlights critical findings, and provides detailed analyses. 

\textbf{Adaptive Teaming without Teammate Modeling.} Fig.~\ref{fig:exp1} and Table~\ref{tab:exp} comprehensively evaluate four baseline methods—SP, PBT, HOLA-Drone (V2) w/o g, and HOLA-Drone (V2)—across four progressively challenging multi-drone pursuit scenarios (4p2e3o, 4p2e1o, 4p2e5o, and 4p3e5o). 
The HOLA-Drone (V2) w/o g variant serves as an ablation study, removing the core hypergraphic game mechanism to assess its impact.
Each scenario is tested against three distinct unseen teammate zoos. Fig.~\ref{fig:exp1} primarily highlights success rates (SUC), while Table~\ref{tab:exp} further details collision counts (COL), average steps taken (AST), and cumulative rewards (REW). 
Together, these metrics demonstrate the efficacy of adaptive teaming methods within the AT-Drone benchmark.

Red dotted lines in Fig.~\ref{fig:exp1} denote approximate upper-bound performances established by best-response (BR) policies specifically trained for each unseen teammate zoo, serving as critical reference points for evaluating adaptive teaming effectiveness. As environmental complexity escalates from the simplest scenario (4p2e1o) to the most complex (4p3e5o), success rates across all evaluated methods decrease, and the performance gap compared to BR policies widens. This clearly illustrates the increasing challenge of effective coordination with unfamiliar teammates.

Overall, HOLA-Drone (V2) consistently demonstrates superior performance across most adaptive teaming scenarios, achieving higher success rates, fewer collisions, and reduced average steps compared to other baseline methods. Notably, the performance advantages of HOLA-Drone (V2) become significantly more pronounced in complex scenarios, underscoring the value of advanced adaptive strategies in dynamic environments involving diverse teammate behaviors.

Interestingly, the simplified ablation model, HOLA-Drone (V2) w/o g—lacking the core hypegraphical-form game module—achieves comparable or even superior performance metrics in simpler scenario 4p2e1o. This observation suggests that while the hierarchical graphical module greatly benefits coordination in challenging environments, it may introduce unnecessary complexity when facing relatively straightforward conditions.

\begin{figure}
    \centering
    \includegraphics[width=\linewidth]{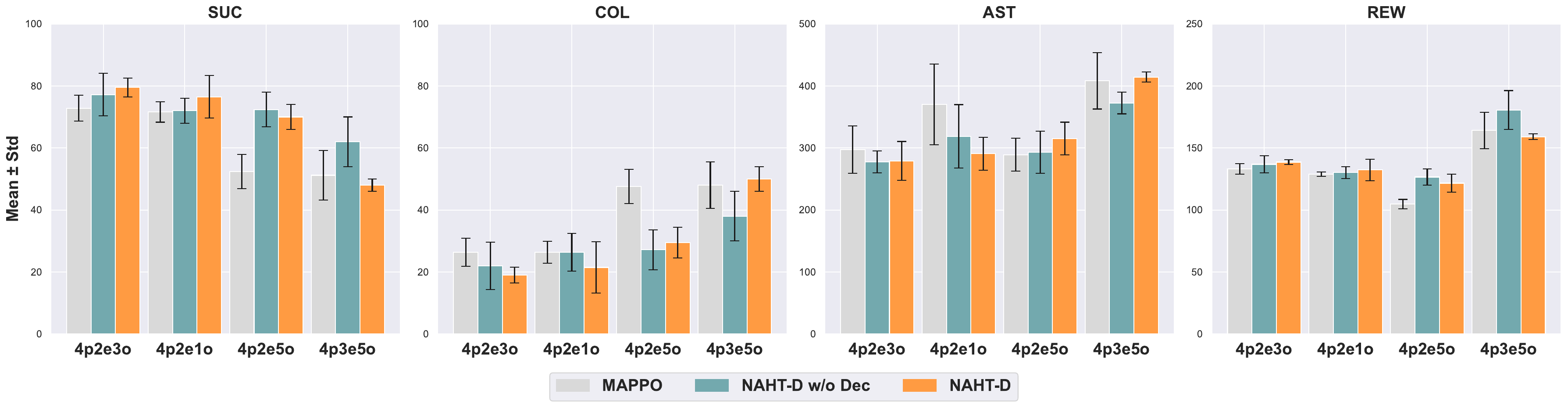}
    \caption{Performance comparison of adaptive teaming with teammate modeling across environments with varying difficulties.}
    \label{fig:exp2}
    \vspace{-5mm}
\end{figure}

\textbf{Adaptive teaming with teammate modelling. } 
Fig.~\ref{fig:exp2} illustrates the comparative performance of MAPPO, NAHT-D w/o Dec, and NAHT-D methods across four multi-drone pursuit scenarios: {4p2e3o}, {4p2e1o}, {4p2e5o}, and {4p3e5o}, explicitly incorporating teammate modeling into the adaptive teaming process. The NAHT-D w/o Dec variant serves as an ablation study, removing the teammate modeling decoder network and the corresponding loss function to assess their impact. 

Across simpler scenarios ({4p2e3o} and {4p2e1o}), NAHT-D achieves marginally higher or comparable success rates (SUC) compared to NAHT-D w/o Dec, with both outperforming MAPPO. However, as scenario complexity increases ({4p2e5o} and {4p3e5o}), NAHT-D w/o Dec demonstrates notably superior performance over NAHT-D, indicating that the additional complexity introduced by the teammate modeling decoder may negatively impact coordination efficiency under highly challenging conditions. 
Regarding collision counts (COL) and average steps (AST), NAHT-D w/o Dec consistently achieves better or comparable results compared to NAHT-D, further suggesting that simpler teammate modeling strategies can offer superior robustness and efficiency in complex adaptive teaming scenarios. These insights are also supported by cumulative rewards (REW), where NAHT-D w/o Dec generally achieves higher or similar values across all settings.

\textbf{Case study and demo videos.} Appendix~\ref{appendix:exp_case} provides a detailed case study demonstrating the real-world deployment of our adaptive teaming approach.
In this case study, ATM learners are paired with unseen drone partners sampled from Zoo 3 and tasked with operating in the most challenging environment~\superhard.
Further demonstration videos can be accessed on our project website at \url{https://sites.google.com/view/at-drone}.

\section{Conclusion}

In this paper, we introduced AT-Drone, a novel multi-robot collaboration benchmark specifically designed to investigate adaptive teaming problems in multi-drone pursuit scenarios. AT-Drone uniquely integrates customizable simulation environments, real-world deployment pipelines leveraging Crazyflie drones and edge computing, a distributed training framework supporting diverse adaptive teaming algorithms, and standardized evaluation protocols.
To facilitate comprehensive evaluation, AT-Drone provides four progressively challenging adaptive teaming environments and a collection of three distinct, unseen drone teammate zoos. Experimental results conducted with Crazyflie drones confirm AT-Drone's effectiveness in driving advancements in adaptive teaming research. Furthermore, real-world drone experiments validate the benchmark's practical feasibility and utility, demonstrating its relevance and applicability for realistic robotic operations.

\section{Limitations} 
While AT-Drone successfully bridges simulation and real-world deployment, the current real-world system remains relatively simple, potentially limiting its ability to capture more intricate scenarios encountered in practical operations. 
Additionally, scalability constraints posed by physical hardware limitations—such as size, payload restrictions inherent to Crazyflie drones, and the current maximum of four pursuers and two evaders within a 3.6m × 5m area—may restrict larger-scale experimentation, making it challenging to study more complex pursuit tasks. Another critical limitation relates to perception; the current setup may not adequately handle complex perception tasks, potentially limiting the drones' adaptability in visually challenging environments. Despite efforts to replicate realistic scenarios, simulation environments may not fully encapsulate all complexities and uncertainties present in real-world conditions. Future research should focus on addressing these limitations by enhancing system complexity, improving perception capabilities, exploring scalable drone hardware alternatives, enhancing simulation fidelity, and developing methods suitable for larger-scale, more complex scenarios.

\section*{Acknowledge}
We sincerely thank Jianghong Wang for constructive feedback during the preparation of this manuscript.

\bibliography{example_paper}

\newpage
\appendix
\onecolumn

\begin{figure}
    \centering
    \includegraphics[width=\linewidth]{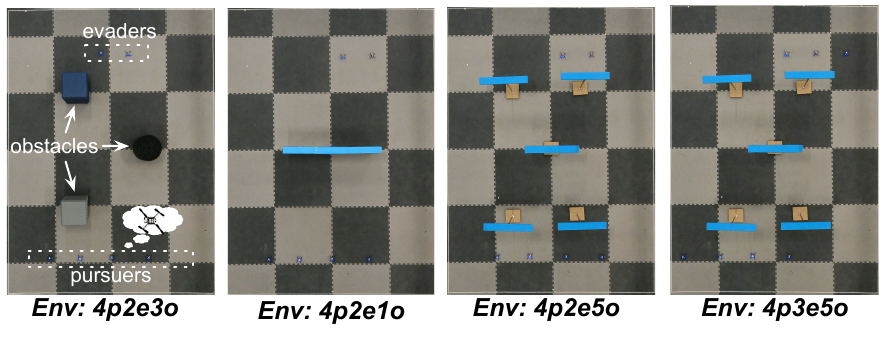}
    \caption{Illustration of four multi-drone pursuit environments in real world. The environments vary in the number of pursuers (p), evaders (e), and obstacles (o), denoted as \texttt{4p2e3o}, \texttt{4p2e1o}, \texttt{4p2e5o}, and \texttt{4p3e5o}. 
    Each setup introduces different levels of complexity, testing the adaptability and coordination capabilities of the agents.}
    \label{fig:app_env}
\end{figure}

\section{Environment Configurator}
\label{appendix:env_config}

The \framework environment configurator allows users to define and modify multi-drone pursuit scenarios through a structured JSON file. Fig.~\ref{fig:env_json} provides an example configuration file that specifies key parameters across three categories: \textit{players}, \textit{site}, and \textit{task}.

\textbf{Players Configuration:}  
This section defines the number and roles of agents in the environment, including the number of pursuers (\texttt{num\_p}), evaders (\texttt{num\_e}), controlled agents (\texttt{num\_ctrl}), and unseen teammates (\texttt{num\_unctrl}). Additional parameters such as random respawn behavior, reception range, and velocity settings further customize agent interactions. The \texttt{unseen\_drones} field allows users to specify different unseen teammate models from the unseen drone zoo.

\textbf{Site Configuration:}  
This section defines the physical properties of the environment, including its boundary dimensions (\texttt{width}, \texttt{height}) and obstacle placements. Obstacles can be configured individually to introduce varying levels of complexity.

\textbf{Task Configuration:}  
This section sets the pursuit task parameters, including the capture range (\texttt{capture\_range}), safety radius (\texttt{safe\_radius}), task duration (\texttt{task\_horizon}), and simulation frame rate (\texttt{fps}). The \texttt{task\_name} field provides a label for different predefined environment scenarios.

This modular configuration enables flexible environment customization, facilitating experiments across diverse multi-drone pursuit scenarios.

\section{Unseen Drone Zoo}
\label{appendix:unseen_zoo}
\textit{Rule-Based Method: Greedy Drone. }  
The Greedy Drone pursues the closest target by continuously aligning its movement with the target’s position. Its state information includes its own position, orientation, distances and angles to teammates and evaders, and proximity to obstacles or walls. When obstacles or other agents enter its evasion range, the Greedy Drone dynamically adjusts its direction to avoid collisions, prioritising immediate objectives over team coordination.

\textit{Traditional Method: VICSEK Drone. }  
Based on the commonly used VICSEK algorithm~\cite{Janosov2017Group,ZhangDACOOP2023,hola-drone}, the VICSEK Drone adopts a bio-inspired approach to mimic swarm-like behaviours. It computes and updates a velocity vector directed towards the evader, optimising the tracking path based on the agent’s current environmental state. To avoid nearby obstacles or agents, the VICSEK Drone applies repulsive forces with varying magnitudes. While the calculated velocity vector includes both magnitude and orientation, only the orientation is implemented in our experiments, making it a scalable and practical teammate model for multi-drone coordination.

\textit{Learning-Based Method: Self-Play Drones.}  
For the learning-based approach, we employ an IPPO-based self-play algorithm, generating diverse drone behaviours by training agents with different random seeds. This approach simulates a wide range of adaptive strategies, introducing stochasticity and complexity to the evaluation process.

\begin{figure}[t]
    \centering
\begin{minipage}{0.5\linewidth}
\begin{lstlisting}[language=json]
{
    "players": {
        "num_p": 4, 
        "num_e": 2, 
        "num_ctrl": 2, 
        "num_unctrl": 2, 
        "random_respawn": True,
        "respawn_region":  {***},
        "reception_range": 2,
        "velocity_p": 0.3,
        "velocity_e": 0.6,
        "unseen_drones": [***]
    },
    "site":{
        "boundary": {
            "width" : 3.6, 
            "height" : 5, 
        },
        "obstacles": {
            "obstacle1":{***}
        },
    },
    "task":{
        "task_name": 4p2e1o,
        "capture_range": 0.2,
        "safe_radius": 0.1,
        "task_horizon": 100, 
        "fps": 10,
    }
}
\end{lstlisting}
\end{minipage}
    \caption{An example of environment configuration file.}
    \label{fig:env_json}
\end{figure}

\section{HOLA-Drone (V2) Algorithm}
\label{appendix:opt}

\begin{figure}[!ht]
    \centering
    \includegraphics[width=0.6\linewidth]{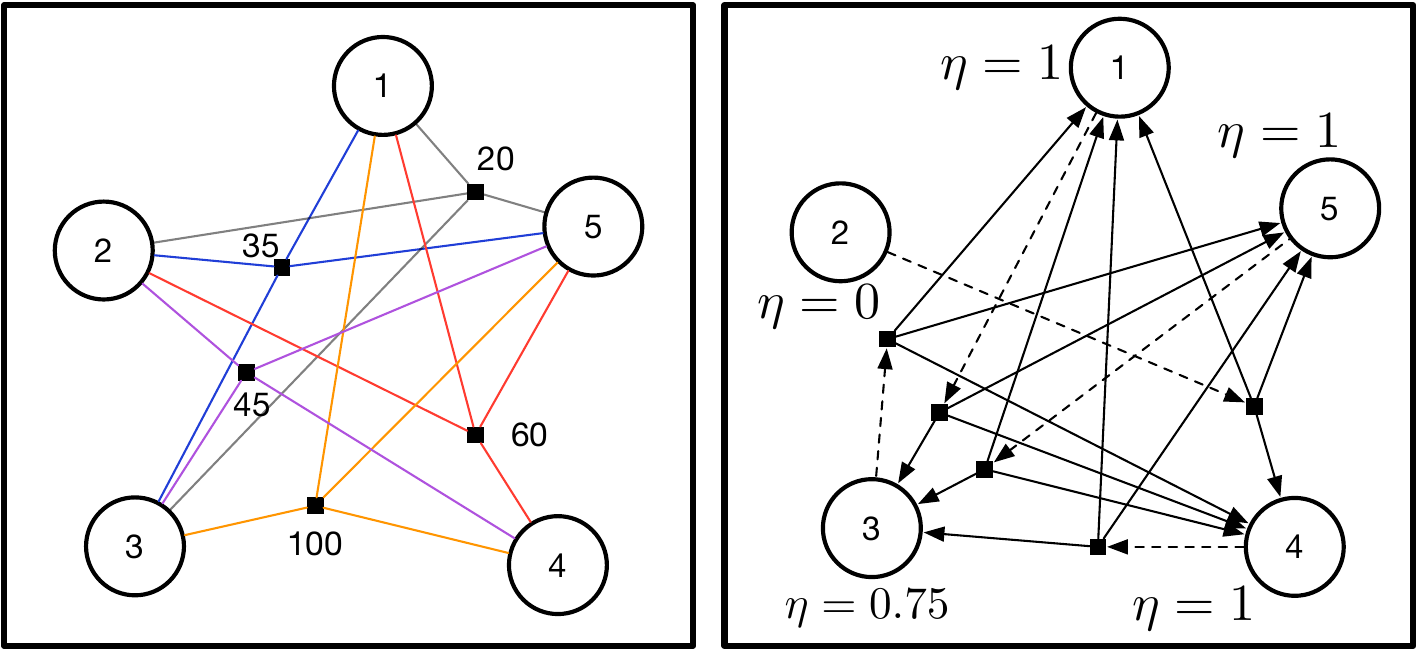}
    \caption{An example of a hypergraph representation (left) and its corresponding preference hypergraph (right) with five strategies in the population.}
    \vspace{-3mm}
    \label{fig:model}
\end{figure}

In this section, we define a population of drone strategies, denoted as $\Pi = \{\pi_1, \pi_2, \cdots, \pi_n\}$. For the task involving $C$ teammates, the interactions within the population $\Pi$ are modeled as a hypergraph $\mathcal{G}$. Formally, the hypergraph is represented by the tuple $(\Pi, \mathcal{E}, \mathbf{w})$, where the node set $\Pi$ represents the strategies, $\mathcal{E}$ is the hyperedge set capturing interaction relationships among teammates, and $\mathbf{w}$ is the weight set representing the corresponding average outcomes. 
The left subfigure of Fig.~\ref{fig:model} illustrates an example of a hypergraph representation with five nodes and a fixed hyperedge length of 4.

Building on the concept of preference hypergraphs~\cite{hola-drone}, we use the preference hypergraph to represent the population and assess the coordination ability of each node. The \textbf{preference hypergraph} $\mathcal{P}\mathcal{G}$ is derived from the hypergraph $\mathcal{G}$, where each node has a direct outgoing hyperedge pointing to the teammates with whom it achieves the highest weight in $\mathcal{G}$. 
Formally, $\mathcal{P}\mathcal{G}$ is defined by the tuple $(\Pi, \mathcal{E}_\mathcal{P})$, where the node set $\Pi$ represents the strategies, and $\mathcal{E}_\mathcal{P}$ denotes the set of outgoing hyperedges. As shown in the right subfigure of Fig.~\ref{fig:model}, the dotted line highlights the outgoing edge. 
For instance, node $2$ has a single outgoing edge $(2, 3, 5, 4)$ because it achieves the highest outcome, i.e., a weight of 45, with those teammates in $\mathcal{G}$, as depicted in the left subfigure.

Intuitively, a node in $\mathcal{P}\mathcal{G}$ with higher cooperative ability will have more incoming hyperedges, as other agents prefer collaborating with it to achieve the highest outcomes. Therefore, we extend the concept of \textbf{preference centrality}~\cite{li2023cooperative} to quantify the cooperative ability of each node. Specifically, for any node $i \in \Pi$, the preference centrality is defined as 
\begin{equation}
    \eta_\Pi(i) = \frac{d_{\mathcal{P}\mathcal{G}}(i)}{d_{\mathcal{G}}(i)},
\end{equation}
where $d_{\mathcal{P}\mathcal{G}}(i)$ denotes the incoming degree of node $i$ in $\mathcal{P}\mathcal{G}$, and $d_{\mathcal{G}}(i)$ represents the degree of node $i$ in $\mathcal{G}$.

\textbf{Max-Min Preference Oracle. }
Building on the basic definition of the preference hypergraph representation, we introduce the concept of Preference Optimality to describe the goal of our training process.

\begin{definition}[Preference Optimal]
    A set of learners \(\mathcal{N}^\star\) of size \(N\) is said to be \textbf{Preference Optimal (PO)} in a hypergraph \(\mathcal{G} = (\Pi, \mathcal{E}, \mathbf{w})\) if, for any set \(\hat{\mathcal{N}} \subseteq \Pi\) of size \(N\), the following condition holds:
    \begin{equation}
        \sum_{s \in \mathcal{N}^\star} \eta_\Pi(s) \geq \sum_{s \in \hat{\mathcal{N}}} \eta_\Pi(s),
    \end{equation}
    where \(\eta_\Pi(s)\) denotes the preference centrality of learner \(s\) in the hypergraph \(\mathcal{G}\).
\end{definition}

While achieving a preference-optimal oracle is desirable, it becomes impractical or prohibitively expensive in large, diverse populations. Therefore, we propose the \textbf{max-min preference oracle}, abbreviated as \textit{oracle} in the rest of this paper, to ensure robust adaptability and maximize cooperative performance under the worst-case teammate scenarios.

\begin{figure}[t]
    \centering
    \includegraphics[width=\linewidth]{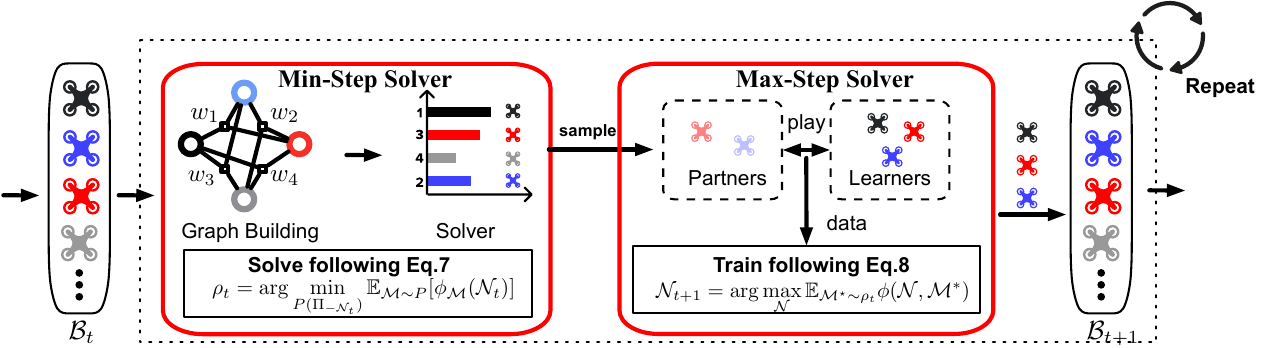}
    \caption{Overview of our proposed HOLA-Drone (V2) algorithm.}
    \label{fig:opt}
\end{figure}
To formalize the objective, we split the strategy population \(\Pi\) into a learner set \(\mathcal{N}\) and a non-learner set \(\Pi_{-\mathcal{N}}\), where \(\Pi_{-\mathcal{N}} \cap \mathcal{N} = \emptyset\) and \(\Pi_{-\mathcal{N}} \cup \mathcal{N} = \Pi\). The objective function \(\phi\) is defined as:
\begin{equation}
    \phi : \underbrace{\mathcal{N} \times \cdots \times \mathcal{N}}_{N \text{ learners}} \times \underbrace{\Pi_{-\mathcal{N}} \times \cdots \times \Pi_{-\mathcal{N}}}_{M \text{ teammates}} \rightarrow \mathbb{R}.
\end{equation}

The max-min preference oracle updates the learner set by solving:
\begin{equation}
\begin{aligned}
    \mathcal{N}^\prime &= oracle(\mathcal{N}, \phi_{\mathcal{M}}(\cdot)) := \arg\max_{\mathcal{N}} \min_{\mathcal{M} \subseteq \Pi_{-\mathcal{N}}} \phi_\mathcal{M}(\mathcal{N}),
\end{aligned}
\end{equation}
where the objective \(\phi_{\mathcal{M}}(\cdot)\) is derived using the extended curry operator~\cite{balduzzi2019open}, originally designed for two-player games, and is expressed as:
\begin{equation}
\begin{aligned}
    &\left[\underbrace{\mathcal{N} \times \cdots \times \mathcal{N}}_{N \text{ learners}} \times \underbrace{\Pi_{-\mathcal{N}} \times \cdots \times \Pi_{-\mathcal{N}}}_{M \text{ teammates}} \rightarrow \mathbb{R} \right] \\
    \rightarrow &\left[
    \underbrace{\Pi_{-\mathcal{N}} \times \cdots \times \Pi_{-\mathcal{N}}}_{M \text{ teammates}} \rightarrow
    \left[
    \underbrace{\mathcal{N} \times \cdots \times \mathcal{N}}_{N \text{ learners}} \rightarrow \mathbb{R}
    \right]
    \right].
\end{aligned}
\end{equation}

Intuitively, \textit{oracle} alternates between two key steps: the minimization step and the maximization step. In the minimization step, the objective is to identify the subset of teammates \( \mathcal{M}^* \subset \Pi_{-\gN} \) that minimizes the performance outcome of the current learner set \( \mathcal{N} \), i.e. the worst partners. This is formulated as:  
\[
\mathcal{M}^* = \arg\min_{\mathcal{M} \subset \Pi_{-\gN}} \phi_\mathcal{M}(\mathcal{N}).
\]  
In the maximization step, the learner set \( \mathcal{N} \) is updated to maximize its performance outcome against the identified subset \( \mathcal{M}^* \). This is defined as:  
\[
\mathcal{N}^* = \arg\max_{\mathcal{N}} \phi(\mathcal{N}, \mathcal{M}^*).
\]  

To achieve robust adaptability and dynamic coordination in multi-agent systems, we integrate the max-min preference oracle into an open-ended learning framework, referred to as the HOLA-Drone (V2) algorithm. 
The HOLA-Drone (V2) algorithm dynamically adjusts the training objective as the population evolves, enabling continuous improvement and effective coordination with unseen partners.
Unlike conventional fixed-objective training, the HOLA-Drone (V2) approach iteratively expands the strategy population \(\Pi\) and refines the learner set \(\mathcal{N}\). At each generation $t$, the framework recalibrates the training objective \(\phi\) based on new extended population $\Pi_t$ to account for the evolving interactions among agents within the population. 

As shown in Fig.~\ref{fig:model}, HOLA-Drone (V2) algorithm consists of two key modules: the min-step solver and the max-step trainer. At each generation \(t\), the updated learner set \(\mathcal{N}_{t}\) from the previous generation \(t-1\) is incorporated into the population \(\Pi_{t-1}\), resulting in an expanded population \(\Pi_{t}\).

\textbf{Min-step Solver. }The role of the min-step solver is to first construct the preference hypergraph representation of the interactions within the updated population \(\Pi_{t}\). Here, we only need to build a subgraph of the entire preference hypergraph in \(\Pi_{t}\), denoted as \(\mathcal{P}\mathcal{G}_t^\prime\). To obtain \(\mathcal{P}\mathcal{G}_t^\prime\), we focus on constructing the hyperedges in the hypergraph \(\mathcal{G}_t^\prime\) that connect to the learner set \(\mathcal{N}_t\).
For instance, if \(\Pi_{t}\) consists of a learner set \(\mathcal{N}_{t}\) of size \(N\) and a non-learner set \(\Pi_{-\mathcal{N}_{t}}\), any hyperedge \(e\) in \(\mathcal{G}_t^\prime\) connects \(N\) nodes from \(\mathcal{N}_{t}\) and all possible \(M\) nodes from \(\Pi_{-\mathcal{N}_{t}}\). The preference hypergraph \(\mathcal{P}\mathcal{G}_t^\prime\) is then derived from \(\mathcal{G}_t^\prime\) by retaining only the outgoing hyperedge with the highest weight for each node.

The min-step solver uses the reciprocal of the preference centrality in \(\mathcal{P}\mathcal{G}_t^\prime\) to evaluate the worst-case partners. To enhance robustness, the solver does not deterministically select the worst-case partners as \(\mathcal{M}^* = \arg\min_{\mathcal{M} \subset \Pi_{-\mathcal{N}}} \phi_\mathcal{M}(\mathcal{N})\). Instead, it outputs a mixed strategy \(\rho_t\), defined as:  
\begin{equation}
    \rho_t = \arg\min_{P(\Pi_{-\mathcal{N}_t})} \mathbb{E}_{\mathcal{M} \sim P}[\phi_\mathcal{M}(\mathcal{N}_t)].\label{eq:final_min}
\end{equation}
In practice, the mixed strategy \(\rho_t\) is obtained by normalizing the reciprocal of the preference centrality, assigning higher probabilities to worse partners.

\textbf{Max-Step Solver. }
Given the mixed strategies \(\rho_t\), the max-step solver iteratively samples the worst-case partners, referred to as the profile, \(\mathcal{M} \sim \rho_t\), from the non-learner set \(\Pi_{-\mathcal{N}_t}\). It simulates interactions between the sampled profile and the learners to generate training data, with the objective of maximizing the reward \(\phi(\mathcal{N}_t, \mathcal{M}) = \mathbb{E}_{\mathcal{N}_t, \mathcal{M}}[R(\tau)]\), as shown in Eq.~\ref{eq:obj}.
The max-step oracle could be rewritten as 
\begin{equation}
    \mathcal{N}_{t+1} = \arg\max_{\mathcal{N}} \mathbb{E}_{\mathcal{M}^\star \sim \rho_t} \phi(\mathcal{N}, \mathcal{M}^*).
    \label{eq:final_max}
\end{equation}

The strategy network for adaptive teaming, supports both AHT and ZSC paradigms. In the AHT paradigm, the network uses a teammate modeling network (\(f\)) to infer teammate types from the observation history (\(\tau_t^{i}\)). These predicted vector, combined with the agent’s observation history (\(\tau_t^i\)), are input into a PPO-based policy network. This policy network includes an Actor Network (\(\pi_\theta\)) for generating the agent’s action (\(a_t^i\)) and a Critic Network (\(V_\pi\)) for evaluating the policy.

In contrast, the ZSC paradigm simplifies the process by directly feeding the agent’s observation history (\(\tau_t^i\)) into the actor and critic networks, bypassing explicit teammate modeling. This approach enables the agent to coordinate with unseen teammates without prior knowledge or additional inference mechanisms.

The max-step solver ultimately generates an approximate best response $\gN_{t+1}$ to the worst-case partners, enhancing the agent’s adaptive coordination capabilities.

\begin{table}[t]
        \centering
        \caption{Implementation hyperparameters of NAHT -D algorithm.}
        \begin{tabular}{@{}cc|cc@{}}
            \toprule
            \textbf{Parameters} & \textbf{Values} & \textbf{Parameters} & \textbf{Values}\\ \midrule
            Batch size & 1024 & Minibatch size & 256\\ 
            Lambda (\(\lambda\)) & 0.99  & Generalized advantage estimation lambda (\(\lambda_{gae}\)) & 0.95\\ 
            Learning rate & 3e-4 & Value loss coefficient($c_1$) & 1 \\ 
            Entropy coefficient(\(\epsilon_{clip}\)) & 0.01 & PPO epoch & 20 \\ 
            Total environment step & 1e6 &  History length & 1 \\ 
            Embedding size & 16 & Hidden size & 128 \\
            \bottomrule
        \end{tabular}
        \label{tab:ATM_params}
    \end{table}

\section{NAHT-D algorithm}
\label{appendix:naht-d}

NAHT-D extends the MAPPO algorithm~\cite{yu2022surprising} by incorporating an additional teammate modeling network \( f \). 
This network generates team encoding vectors to represent the behavioral characteristics of unseen teammates, improving coordination in multi-drone pursuit.
The modeling network \( f \) processes three types of inputs: (1) observed evader states history, (2) self-observed states history, and (3) relative positions history between agents. These inputs are transformed using independent fully connected layers, aggregated through weighted averaging, and combined into a unified team representation. The final embedding is a fixed-dimensional vector, integrated into the actor network of MAPPO to enhance decision-making in adaptive teaming scenarios.
The details of hyperparameters are listed in Table~\ref{tab:ATM_params}.

\begin{figure}[t]
    \centering
    \includegraphics[width=0.85\linewidth]{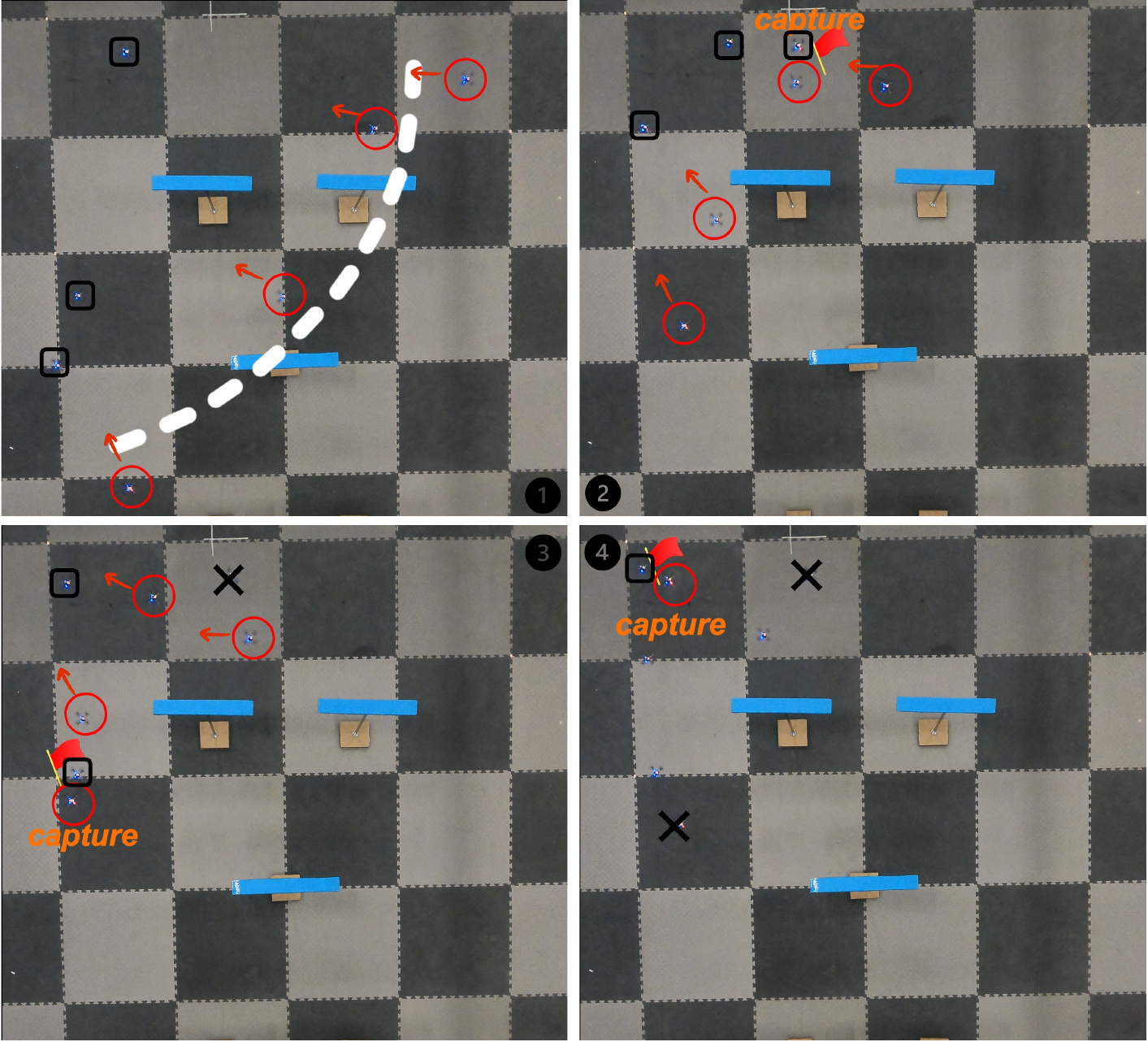}
    \caption{\textbf{Case Study:} This example demonstrates the capture strategy executed by NAHT-D learners and unseen drone partners from unseen zoo 3 in the superhard environment \superhard. The red circles denote pursuers, and the black squares represent evaders. 
    In this scenario, four pursuers collaboratively surround all three evader (1), two pursuers capture one of evaders while other two pursuers continuously tighten their formation (2), and rest of two evaders are then successfully captured one by one (3 \& 4)}
    \label{fig:case_study}
\vspace{-3mm}
\end{figure}

\section{Case Study}
\label{appendix:exp_case}
To further illustrate the effectiveness of our adaptive teaming approach, we present a case study in the \texttt{4p3e5o} environment, categorized as superhard due to its high complexity, featuring four pursuers, three evaders, and five obstacles. 
The unseen teammates in this scenario are sampled from Unseen Zoo 3, which consists entirely of PPO-based self-play policies trained at two different skill levels, introducing high adaptability and unpredictability.

This case study demonstrates how the NAHT-D learners effectively coordinate with their unseen drone partners to execute a multi-stage capture strategy. Fig.~\ref{fig:case_study} illustrates key frames from the scenario. In Frame 1, four pursuers initiate a collaborative approach, positioning themselves strategically to encircle all three evaders while maintaining an adaptive formation. In Frame 2, two pursuers successfully capture one of the evaders while the other two tighten their formation, preventing the remaining evaders from escaping. In Frames 3 \& 4, the remaining two evaders are captured one by one as the pursuers continue refining their positioning and coordination, effectively closing all escape routes.

\end{document}